\documentclass[runningheads]{llncs}
\usepackage[T1]{fontenc}
\usepackage{graphicx}
\usepackage{marvosym}
\usepackage{color}
\usepackage{url}
\usepackage{amsmath}
\usepackage{amssymb}
\usepackage{multirow}
\usepackage{multicol}
\usepackage{makecell}
\usepackage{booktabs}
\usepackage{colortbl}
\usepackage{sectsty}

\begin{document}
\title{MediCLIP: Adapting CLIP for Few-shot Medical Image Anomaly Detection}
%
%
\author{
Ximiao Zhang\inst{1} \and
Min Xu\inst{1}\textsuperscript{*} \and
Dehui Qiu\inst{1} \and
Ruixin Yan\inst{2} \and
Ning Lang\inst{2} \and
Xiuzhuang Zhou\inst{3}\textsuperscript{*}
}

\authorrunning{Ximaio Zhang et al.}
%

\institute{
College of Information and Engineering, Capital Normal University, Beijing, China \and
Department of Radiology, Peking University Third Hospital, Beijing, China \and
School of Artificial Intelligence, Beijing University of Posts and Telecommunications, Beijing, China 
}

\maketitle              

\begin{abstract}

In the field of medical decision-making, precise anomaly detection in medical imaging plays a pivotal role in aiding clinicians. However, previous work is reliant on large-scale datasets for training anomaly detection models, which increases the development cost. This paper first focuses on the task of medical image anomaly detection in the few-shot setting, which is critically significant for the medical field where data collection and annotation are both very expensive. We propose an innovative approach, MediCLIP, which adapts the CLIP model to few-shot medical image anomaly detection through self-supervised fine-tuning. Although CLIP, as a vision-language model, demonstrates outstanding zero-/few-shot performance on various downstream tasks, it still falls short in the anomaly detection of medical images. To address this, we design a series of medical image anomaly synthesis tasks to simulate common disease patterns in medical imaging, transferring the powerful generalization capabilities of CLIP to the task of medical image anomaly detection. When only few-shot normal medical images are provided, MediCLIP achieves state-of-the-art performance in anomaly detection and location compared to other methods. Extensive experiments on three distinct medical anomaly detection tasks have demonstrated the superiority of our approach. The code is available at \url{https://github.com/cnulab/MediCLIP}.

\end{abstract}
\footnotetext{\scriptsize{* Corresponding author. Email: \email{xumin@cnu.edu.cn}, \email{xiuzhuang.zhou@bupt.edu.cn}.}}

\section{Introduction}

The medical image anomaly detection task involves distinguishing between normal and anomalous images with various lesion areas, and identifying the locations of these lesion areas. This offers significant insights for medical professionals in making clinical decisions, substantially reducing the risk associated with decision-making and enhancing the efficiency of medical professionals. Due to the scarcity of lesion images, research in this field is focused on unsupervised anomaly detection settings, where models are trained solely on normal images, and any subtle deviation from the normal pattern is identified as an anomaly. Recently, significant progress has been made in medical image anomaly detection~\cite{wolleb2022diffusion,wyatt2022anoddpm,cai2022dual,xiang2023squid}, achieving comparable or even better anomaly detection performance than professional clinicians in certain detection tasks~\cite{cao2023large}. However, these methods utilize thousands or more medical images to train their anomaly detection models, increasing the developmental cost and hindering rapid deployment. In this paper, we focus on a more challenging task, namely, the few-shot medical image anomaly detection task. For each medical anomaly detection task, we provide only a few-shot normal images for model training, without any available anomaly images and pixel-level labels. This setup significantly reduces the development cost of medical anomaly detection models and avoids the expensive process of medical data collection and annotation.

Parallel to this, CLIP~\cite{radford2021learning} demonstrates powerful generalization capabilities by mapping natural images and raw texts into a unified representational space, which can be easily applied to a variety of downstream tasks. MedCLIP~\cite{wang2022medclip} extends the capabilities of CLIP~\cite{radford2021learning} for medical image classification by training on unpaired medical images and text. Despite this advancement, the method falls short in identifying and localizing lesion areas. Recent studies~\cite{zhou2023anomalyclip,chen2023clip,chen2023zero,huang2024adapting} have applied CLIP~\cite{radford2021learning} for zero-/few-shot anomaly detection with impressive performance. However, these methods require an auxiliary dataset with real anomaly images and corresponding pixel-level anomaly labels for model training, which is often challenging to obtain in the medical field. To address this limitation, we propose MediCLIP, which employs synthetic anomaly images for model training instead of relying on an auxiliary dataset. These synthetic images, crafted through a variety of carefully designed anomaly synthesis tasks, effectively emulate diverse abnormal patterns observed in medical imaging. In our approach, we leverage learnable prompts~\cite{zhou2022learning} to avoid the complex design of artificial text prompts, ensuring that the text embeddings can effectively generalize within medical images. Additionally, we use adapters to align the intermediate layer features from the CLIP~\cite{radford2021learning} vision encoder with learnable text features, enabling multi-scale lesion localization capabilities.

We validate the effectiveness of MediCLIP on three distinct medical imaging datasets: the chest X-ray image dataset CheXpert~\cite{irvin2019chexpert}, the brain MRI image dataset BrainMRI~\cite{nav2019brainmri}, and the breast ultrasound image dataset BUSI~\cite{al2020dataset}. Under the few-shot learning setting, MediCLIP demonstrates approximately a 10\% improvement in performance compared to existing state-of-the-art methods. Notably, on the most challenging CheXpert~\cite{irvin2019chexpert} dataset, MediCLIP achieved 94\% of the performance level of the advanced full-shot medical image anomaly detection method, SQUID~\cite{xiang2023squid}, while using less than 1\% of the training images. Additionally, we have also discovered that MediCLIP has an impressive zero-shot generalization capability. The anomaly detection ability of MediCLIP acquired by training in a medical anomaly detection task can be transferred to other medical anomaly detection tasks. This indicates that MediCLIP has the potential to serve as a unified model for medical anomaly detection. Through extensive experimentation, we have demonstrated that MediCLIP has both low cost and high performance, showcasing its capabilities in assisting professional doctors with real medical decision-making.

\begin{figure}[t]
  \centering
\includegraphics[width=\textwidth]{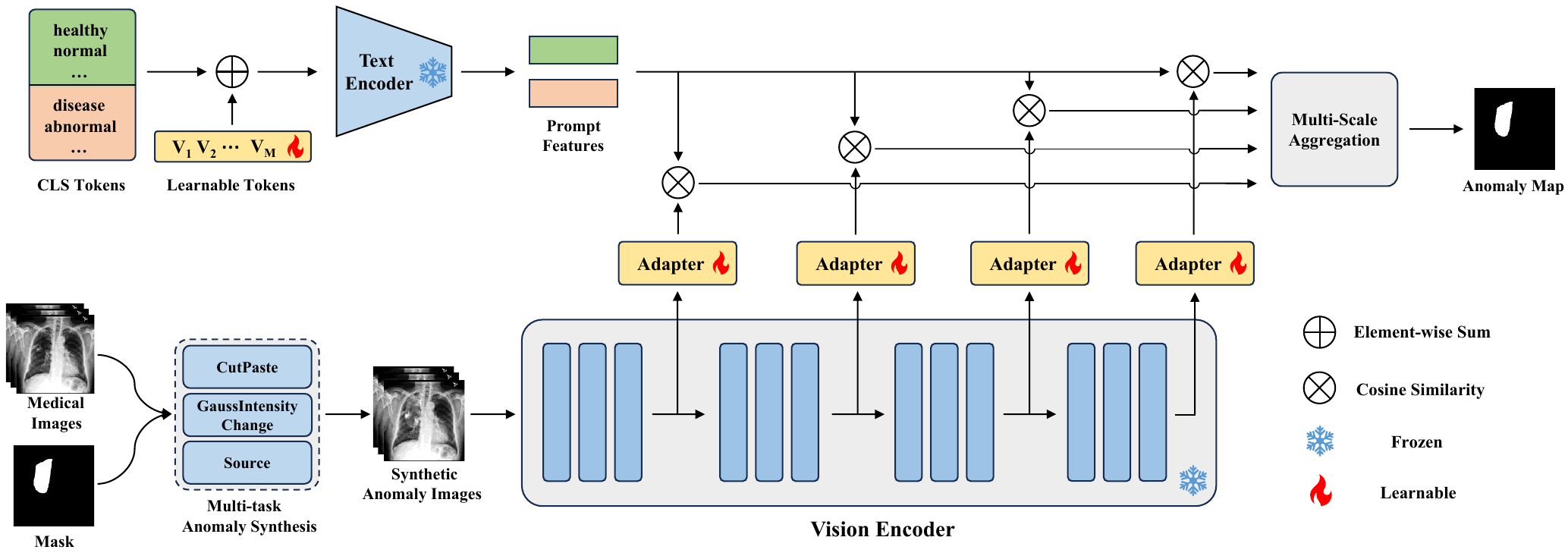}
\caption{The overall pipeline of our proposed MediCLIP framework.}
\label{fig:fig1}
\end{figure}

\section{Method}

In this section, we first present the problem definition of few-shot anomaly detection. Then we introduce the learnable prompts and adapters in MediCLIP. Finally, we describe our medical image anomaly synthesis strategy. The overall pipeline of MediCLIP is illustrated in Fig. \ref{fig:fig1}.

\subsection{Problem Definition}

We provide a formal definition for the few-shot anomaly detection task. According to the few-shot learning setting, for an $n$-way and $k$-shot episode, a support set $D=\bigcup_{i=1}^{n}T_{i}$ contains normal images from $n$ tasks, where $T_i$ includes $k$ normal images from the $i$th task. During the test phase, query images from $n$ tasks are provided, and the model is trained to identify whether these query images are anomalous and to indicate the location of the anomalous regions. Due to the significant differences of various medical anomaly detection tasks, and in order to meet the actual needs of the medical system, in this paper, we set $n=1$, which means one model for one task.

\subsection{Prompt Learning}

In CLIP~\cite{radford2021learning}, commonly used prompts like \texttt{`A photo of a [CLS]'} mainly describe the overall semantic information of the nature image, and they struggle to capture the subtle details in medical imaging. Therefore, we use learnable prompts~\cite{zhou2022learning} instead of manually designed prompts for anomaly detection, ensuring that the text embeddings can generalize effectively in medical images while avoiding the complexity of prompt engineering. Specifically, we adopt the following prompt format:
\begin{equation}
\label{eq:equ1}
p=[V_1][V_2]...[V_M][CLS]
\end{equation}
where $[V]$ and $[CLS]$ represent the learnable word embeddings and the non-learnable class embeddings in the prompt template, and $M$ denotes the count of the learnable tokens. For class tokens, we use medically relevant terms such as \texttt{[healthy]} and \texttt{[normal]} for normal cases, and \texttt{[disease]} for anomalies. The complete list of class tokens is provided in the Appendix. Therefore, for normal and anomalous cases, we have $\mathcal{P}_{n}=\{p_{n_1}, p_{n_2},..., p_{n_I}\}$ and $\mathcal{P}_{a}=\{p_{a_1}, p_{a_2},..., p_{a_E}\}$, where $\mathcal{P}_{n}$ and $\mathcal{P}_{a}$ denote the prompt sets for normal and anomalous cases, and $I$ and $E$ represent the count of class tokens contained within $\mathcal{P}_{n}$ and $\mathcal{P}_{a}$. We define the text encoder of CLIP as $F(\cdot)$, where $F(\cdot)$ maps the prompt $p$ to its feature representation $F(p) \in R^{C}$, where $C$ signifies the dimension of prompt feature. The mean feature representations $f_n$ and $f_a$ for the prompt sets $\mathcal{P}_{n}$ and $\mathcal{P}_{a}$ are calculated respectively by $f_{n}=\dfrac{1}{I}\sum_{i=1}^{I}F(p_{n_i})$ and $f_{a}=\dfrac{1}{E}\sum_{i=1}^{E}F(p_{a_i})$.

\subsection{Adapting CLIP for Anomaly Detection}

The vanilla CLIP model~\cite{radford2021learning} is designed for zero-/few-shot image-text classification and is not directly applicable to anomaly detection and localization. In this section, we introduce adapters to the vanilla CLIP model~\cite{radford2021learning}, adapting it for few-shot anomaly detection through a small number of learnable parameters. Specifically, for a medical image $X \in R^{H \times W \times 3}$ from the support set $D$, we randomly sample an anomaly region mask $Y \in R^{H \times W}$ and generate the corresponding synthetic anomaly image $\hat{X}=\Psi(X,Y)$ through a series of anomaly synthesis tasks, where $\Psi(\cdot)$ represents the anomaly synthesis function. The vision encoder of CLIP ~\cite{radford2021learning}, denoted as, $G_{j}(\cdot)$, which extracts the $j$th layer feature of the image, represented as $G_{j}(\hat{X}) \in R^{H_{j} \times W_{j} \times C_{j}}$, where $H_{j}$, $W_{j}$, and $C_{j}$ respectively represent the height, width, and number of channels of the $j$th layer’s features. Consequently, for image $\hat{X}$, we obtain a multi-scale visual feature set $\{ G_{1}(\hat{X}),G_{2}(\hat{X}),...,G_{J}(\hat{X}) \}$. For each intermediate layer feature $G_{j}(\hat{X})$, we use an adapter $\phi_{j}(\cdot)$ to map it to the same number of channels as the prompt features $f_n$ and $f_a$, resulting in $g_{j}=\phi_{j}(G_{j}(\hat{X}))$, where $g_{j} \in R^{H_{j} \times W_{j} \times C}$. We then calculate the similarity of $g_j$ with $f_n$ and $f_a$ at the spatial location $(h,w)$:
\begin{gather}
\footnotesize
\begin{align}
S_{n_{j}}^{(h,w)}=\frac{exp(<g_{j}^{(h,w)},f_n>/\tau)}{\sum_{f \in \{f_n,f_a\}}exp(<g_{j}^{(h,w)},f>/\tau) },
S_{a_{j}}^{(h,w)}=\frac{exp(<g_{j}^{(h,w)},f_a>/\tau)}{\sum_{f \in \{f_n,f_a\}}exp(<g_{j}^{(h,w)},f>/\tau) }
\end{align}
\end{gather}
where $<\cdot,\cdot>$ represents the cosine similarity, $\tau$ is the temperature parameter, and $h < H_j$, $ w< W_ j$, $h,w \in \mathbb{N}$. For each visual feature $g_j$ projected by adapter, we can obtain similarity matrices $S_{n_j},S_{a_j} \in R^{H_{j} \times W_{j}} $. Then, we perform an aggregation operation on the multi-scale similarity matrix sets $\{ S_{n_1},S_{n_2},...,S_{n_J}\}$ and $\{ S_{a_1},S_{a_2},...,S_{a_J}\}$. We upsample them to the same spatial resolution of $H \times W$, and compute their average to obtain $S_n,S_a \in R^{H \times W}$. To optimize the parameters in learnable prompts and adapters, we define the loss function as:
\begin{equation}
\label{eq:equ3}
\mathcal{L}=\mathcal{L}_{focal}([S_n,S_a],Y)+\mathcal{L}_{dice}(S_a,Y)
\end{equation}
where $\mathcal{L}_{focal}(\cdot,\cdot)$ and $\mathcal{L}_{dice}(\cdot,\cdot)$ respectively represent Focal loss~\cite{lin2017focal} and Dice loss~\cite{li2019dice}, and $[\cdot,\cdot]$ denotes concatenation along the channel. During the inference stage, we remove anomaly synthesis tasks, using the query image as $\hat{X}$. We employ $S_a$ as the pixel-level anomaly map, and the maximum value in $S_a$ is used as the image-level anomaly score.

\subsection{Multi-task Anomaly Synthesis}

\begin{figure}[t]
  \centering
\includegraphics[width=\textwidth]{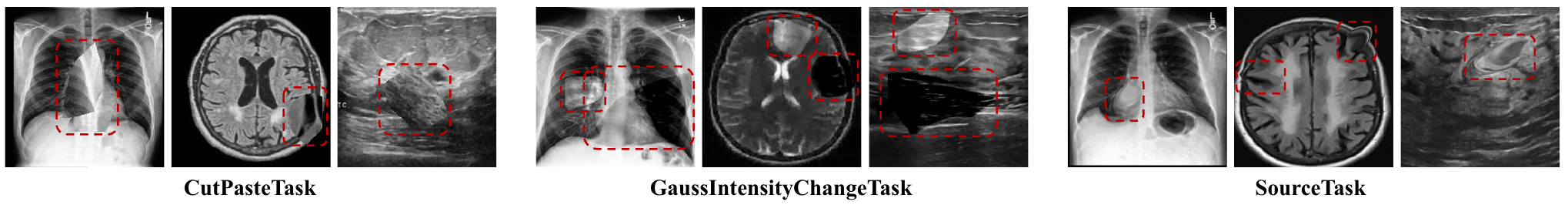}
\caption{Synthetic anomaly image examples from diverse tasks.}
\label{fig:fig2}
\end{figure}

Multi-task anomaly synthesis generates the anomaly image $\hat{X}$ based on a given source image $X$ and a target mask $Y$. This paper introduce three types of anomaly synthesis tasks: CutPaste, GaussIntensityChange, and Source, corresponding to image blending, intensity variation, and deformation, respectively. Some examples are presented in Fig. \ref{fig:fig2}.

\textbf{CutPaste}~\cite{li2021cutpaste,schluter2022natural,baugh2023many} randomly selects an image patch and pastes it to target location. We employ the enhanced version from~\cite{schluter2022natural,baugh2023many} that utilizes Poisson image editing~\cite{perez2023poisson} for seamless image blending. The CutPaste task can simulate misplacement-like anomalies in medical imaging, such as fractures.

\textbf{GaussIntensityChange} simulates density variations in medical imaging by altering pixel intensities. Specifically, we sample noise $\sigma$ from a standard Gaussian distribution $\sigma \sim \mathcal N(0,\textbf{I})$, and then binarize $\sigma$ based on a threshold value of 0. Subsequently, the binarized noise is processed through a Gaussian filter to obtain $\hat{\sigma}$. The final anomaly image is defined as:
\begin{equation}
\label{eq:equ4}
\hat{X}=X \odot (1-Y) + (X+\gamma\hat{\sigma}) \odot Y
\end{equation}
where $\odot$ denotes the element-wise multiplication operation, and $\gamma$ is the intensity variation factor, when $\gamma>0$, intensity increases, and when $\gamma<0$, intensity decreases. The task is particularly effective in mimicking the density variations seen in tumors or cysts within radiographic images.

\textbf{Source}~\cite{baugh2023many} pushes all points in the mask $Y$ away from a center $c$. We define the radius of $Y$ in a certain direction as $r$. For the source position $l$ in this direction, we compute the corresponding target position $\hat{l}$:
\begin{equation}
\label{eq:equ5}
\hat{l}=c + r\frac{l-c}{\|l-c \|_{2}}\left(\frac{\|l-c \|_{2}}{r}\right)^{\alpha},\:c \in \mathbb{L}_{Y},\:\forall\,l \in \mathbb{L}_{Y}
\end{equation}
where $\alpha$ controls the intensity of repulsion from the center,  with a larger $\alpha$ resulting in stronger repulsion. Finally, the pixel at the source location is replaced with the pixel at the target location, resulting in $\hat{x}_{l}=x_{\hat{l}}$. The Source task simulates proliferative anomalies found in medical imaging, such as cardiomegaly.

\section{Experiments}
\subsection{Experimental Setup}

\begin{table}[t]
  \renewcommand\arraystretch{1.0}
  \centering
  \caption{Performance comparison in anomaly detection: MediCLIP versus other methods under few-shot settings, using Image-AUROC (\%) as the evaluation metric.}
    \resizebox{\linewidth}{!}{
    \footnotesize	
    \begin{tabular}{c|c|c|c|c|c|c|c|c|c}
    \bottomrule
    \multicolumn{1}{c|}{\;\;Datasets\;\;} & \multicolumn{1}{c|}{$\;\;k\;\;$} & PatchCore  & SimpleNet  & {\;\;SQUID\;\;}  & AnoDDPM    & CutPaste   & PatchSVDD  & {\; RegAD \;}  & MediCLIP \\
    \hline
    \multicolumn{1}{c|}{\multirow{4}[2]{*}{CheXpert}} 
      & 4     & 60.0{\scriptsize $\pm$1.8} & 63.5{\scriptsize$\pm$2.4} & 66.0{\scriptsize$\pm$1.9} & 55.8{\scriptsize$\pm$1.3}   & 63.6{\scriptsize$\pm$2.8} & 57.5{\scriptsize$\pm$1.1} & 59.8{\scriptsize$\pm$1.4}  & \textbf{70.1{\scriptsize$\pm$1.6}} \\
     & 8     & 59.9{\scriptsize $\pm$2.2} & 63.8{\scriptsize$\pm$0.7} & 66.6{\scriptsize$\pm$3.9} & 57.3{\scriptsize$\pm$3.5}   & 65.0{\scriptsize$\pm$2.9} & 61.3{\scriptsize$\pm$1.5} & 62.0{\scriptsize$\pm$3.0}  & \textbf{70.8{\scriptsize$\pm$1.5}} \\
     & 16    & 64.6{\scriptsize $\pm$1.6} & 65.6{\scriptsize$\pm$3.7} & 67.6{\scriptsize$\pm$4.7} & 59.7{\scriptsize$\pm$2.8}   & 65.2{\scriptsize$\pm$3.2} & 62.7{\scriptsize$\pm$2.3} & 60.2{\scriptsize$\pm$1.7}  & \textbf{72.5{\scriptsize$\pm$1.7}} \\
     & 32    & 64.3{\scriptsize $\pm$1.2} & 67.6{\scriptsize$\pm$1.1} & 69.4{\scriptsize$\pm$2.2} & 66.9{\scriptsize$\pm$3.2}   & 71.4{\scriptsize$\pm$2.0} & 61.7{\scriptsize$\pm$1.4} & 65.6{\scriptsize$\pm$1.4}  & \textbf{73.3{\scriptsize$\pm$1.3}} \\
    \hline
    \multicolumn{1}{c|}{\multirow{4}[2]{*}{BrainMRI}} 
        & 4     & 70.8{\scriptsize $\pm$3.5} & 77.0{\scriptsize$\pm$3.0} & 69.5{\scriptsize$\pm$4.4} & 67.2{\scriptsize$\pm$1.8}   & 72.3{\scriptsize$\pm$1.8} & 78.4{\scriptsize$\pm$1.4} & 67.5{\scriptsize$\pm$1.0}  & \textbf{94.1{\scriptsize$\pm$0.7}} \\
        & 8     & 75.9{\scriptsize $\pm$2.9} & 79.3{\scriptsize$\pm$2.9} & 75.8{\scriptsize$\pm$1.9} & 73.0{\scriptsize$\pm$3.6}   & 77.0{\scriptsize$\pm$4.9} & 79.5{\scriptsize$\pm$0.8} & 75.1{\scriptsize$\pm$2.8}  & \textbf{94.9{\scriptsize$\pm$0.4}} \\
        & 16    & 79.5{\scriptsize $\pm$1.9} & 81.7{\scriptsize$\pm$4.4} & 77.2{\scriptsize$\pm$0.6} & 77.2{\scriptsize$\pm$2.5}   & 80.2{\scriptsize$\pm$2.7} & 78.3{\scriptsize$\pm$3.0} & 82.4{\scriptsize$\pm$1.6}  & \textbf{94.8{\scriptsize$\pm$0.5}} \\
        & 32    & 81.7{\scriptsize $\pm$0.2} & 82.6{\scriptsize$\pm$0.9} & 79.3{\scriptsize$\pm$1.7} & 80.0{\scriptsize$\pm$2.1}   & 79.0{\scriptsize$\pm$4.3} & 78.8{\scriptsize$\pm$1.4} & 83.0{\scriptsize$\pm$2.4}  & \textbf{95.4{\scriptsize$\pm$0.4}} \\
    \hline
    \multicolumn{1}{c|}{\multirow{4}[2]{*}{BUSI}} 
          & 4     & 81.7{\scriptsize $\pm$1.3} & 76.1{\scriptsize$\pm$2.5} & 55.4{\scriptsize$\pm$2.2} & 69.3{\scriptsize$\pm$1.3}   & 75.7{\scriptsize$\pm$1.4} & 69.2{\scriptsize$\pm$1.1} & 74.7{\scriptsize$\pm$5.0}  & \textbf{88.5{\scriptsize$\pm$0.6}} \\
          & 8     & 82.6{\scriptsize $\pm$2.1} & 80.4{\scriptsize$\pm$4.8} & 58.7{\scriptsize$\pm$1.6} & 72.6{\scriptsize$\pm$4.6}   & 78.2{\scriptsize$\pm$2.1} & 70.6{\scriptsize$\pm$2.4} & 75.8{\scriptsize$\pm$2.7}  & \textbf{89.0{\scriptsize$\pm$0.3}} \\
          & 16    & 87.6{\scriptsize $\pm$1.0} & 84.1{\scriptsize$\pm$2.7} & 64.8{\scriptsize$\pm$4.5} & 74.4{\scriptsize$\pm$2.9}   & 80.0{\scriptsize$\pm$0.9} & 73.7{\scriptsize$\pm$6.0} & 76.5{\scriptsize$\pm$1.2}  & \textbf{90.8{\scriptsize$\pm$0.4}} \\
          & 32    & 88.9{\scriptsize $\pm$0.1} & 88.0{\scriptsize$\pm$0.3} & 67.8{\scriptsize$\pm$0.4} & 76.2{\scriptsize$\pm$4.0}   & 80.2{\scriptsize$\pm$1.7} & 77.9{\scriptsize$\pm$1.6} & 78.6{\scriptsize$\pm$1.2}  & \textbf{91.0{\scriptsize$\pm$0.3}} \\
    \toprule
    \end{tabular}
    }
  \label{tab:table1}
\end{table}

\textbf{Datasets.} We conducted extensive experiments on three datasets, including the Stanford CheXpert dataset~\cite{irvin2019chexpert}, the BrainMRI dataset~\cite{nav2019brainmri}, and the BUSI dataset~\cite{al2020dataset}. The CheXpert dataset~\cite{irvin2019chexpert} comprises chest X-ray images from clinical patients at Stanford Hospital, encompassing 12 different diseases (such as Cardiomegaly, Edema, Pleural Effusion, etc.). The BrainMRI dataset~\cite{al2020dataset} contains 2D human brain MRI images, including both normal and tumor-affected cases. The BUSI dataset~\cite{al2020dataset} consists of breast ultrasound images from female patients aged 25 to 75. These images are classified into normal, benign, and malignant categories, with pixel-level annotations marking disease locations. We treated all disease images as anomalies. For each dataset, we provided $k=\{4,8,16,32\}$ normal images for model training. During the testing phase, the CheXpert~\cite{irvin2019chexpert}, BrainMRI~\cite{nav2019brainmri}, and BUSI~\cite{al2020dataset} datasets contain 250, 65, and 101 normal images, respectively, along with 250, 155, and 647 anomaly images.

\begin{table}[t]
  \renewcommand\arraystretch{1.0}
  \centering
  \caption{Performance comparison in anomaly localization: MediCLIP versus other methods on the BUSI dataset~\cite{al2020dataset}, using Pixel-AUROC (\%) as the evaluation metric.}
    \resizebox{\linewidth}{!}{
    \scriptsize	
    \begin{tabular}{c|c|c|c|c|c|c|c}
        \bottomrule
        \multicolumn{1}{c|}{$\;\;k\;\;$} & PatchCore & SimpleNet & {\;\;SQUID\;\;}  & AnoDDPM & PatchSVDD & {\; RegAD \;} & MediCLIP \\
        \hline
        4     & 78.5{\tiny$\pm$2.6} & 73.5{\tiny$\pm$4.9} & 56.8{\tiny$\pm$5.1} & 68.3{\tiny$\pm$1.9} & 74.4{\tiny$\pm$0.9} & 71.9{\tiny$\pm$3.1} & \textbf{88.0{\tiny$\pm$0.3}} \\
        8     & 78.1{\tiny$\pm$1.3} & 77.6{\tiny$\pm$3.4} & 56.8{\tiny$\pm$1.2} & 74.7{\tiny$\pm$4.3} & 73.6{\tiny$\pm$3.4} & 76.1{\tiny$\pm$2.6} & \textbf{88.6{\tiny$\pm$1.0}} \\
        16    & 80.2{\tiny$\pm$0.8} & 78.2{\tiny$\pm$1.5} & 65.5{\tiny$\pm$2.8} & 79.5{\tiny$\pm$1.4} & 78.3{\tiny$\pm$2.2} & 74.6{\tiny$\pm$1.0} & \textbf{89.9{\tiny$\pm$0.9}} \\
        32    & 80.3{\tiny$\pm$0.1} & 79.8{\tiny$\pm$0.4} & 68.4{\tiny$\pm$2.2} & 79.1{\tiny$\pm$1.3} & 77.6{\tiny$\pm$3.4} & 76.3{\tiny$\pm$0.4} & \textbf{90.1{\tiny$\pm$1.0}} \\
        \toprule
        \end{tabular}%
        }
  \label{tab:table2}
\end{table}

\begin{figure*}[t]
\begin{minipage}{0.39\textwidth}
		\centering
		\makeatletter\def\@captype{table}\makeatother
        \renewcommand\arraystretch{1.0}
        \caption{\footnotesize Zero-shot anomaly detection performance of MediCLIP with support set size $k = 16$.}
        \vspace{4pt}
        \resizebox{\linewidth}{!}{
        \begin{tabular}{c|c|c}
        \bottomrule
        \multicolumn{1}{c|}{\;Train Task\;} & {\;\:Test Task\;\:} & Image-AUROC (\%) \\
        \hline
        \multicolumn{1}{c|}{\multirow{3}[2]{*}{CheXpert}} & \cellcolor[rgb]{ .851,  .851,  .851} CheXpert & \cellcolor [rgb]{ .851,  .851,  .851}72.5{\scriptsize$\pm$1.7} \\
              & BrainMRI & 92.1{\scriptsize$\pm$1.2} \\
              & BUSI  & 88.8{\scriptsize$\pm$1.0} \\
        \hline
        \multicolumn{1}{c|}{\multirow{3}[2]{*}{BrainMRI}} & CheXpert & 67.7{\scriptsize$\pm$0.8} \\
              & \cellcolor[rgb]{ .851,  .851,  .851}BrainMRI & \cellcolor[rgb]{ .851,  .851,  .851}94.8{\scriptsize$\pm$0.5} \\
              & BUSI  & 84.8{\scriptsize$\pm$1.1} \\
        \hline
        \multicolumn{1}{c|}{\multirow{3}[2]{*}{BUSI}} & CheXpert & 70.7{\scriptsize$\pm$0.8} \\
              & BrainMRI & 91.3{\scriptsize$\pm$1.0} \\
              & \cellcolor[rgb]{ .851,  .851,  .851}BUSI & \cellcolor[rgb]{ .851,  .851,  .851}90.8{\scriptsize$\pm$0.4} \\
        \toprule
        \end{tabular}%
        }
        \label{tab:table3}
\end{minipage}\quad
\begin{minipage}{0.578\textwidth}
        \centering
        \makeatletter\def\@captype{figure}\makeatother
        \includegraphics[width=\linewidth]{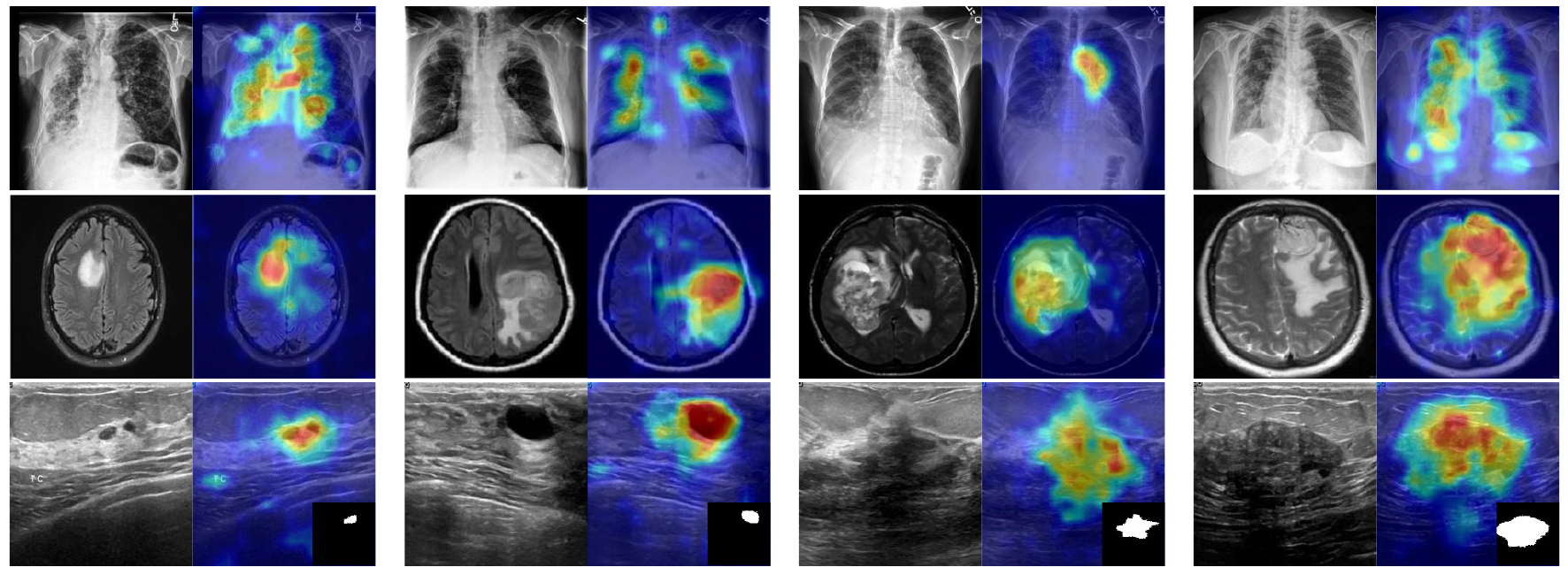}
        \caption{\footnotesize	 Qualitative results of MediCLIP. Each group contains the anomaly image and the predicted anomaly map.}
    \label{fig:fig3}
\end{minipage}
\end{figure*}

\noindent \textbf{Implementation Details.} We used the ViT-L/14~\cite{radford2021learning} as the pre-trained CLIP model, retaining the features of the 12th, 18th, and 24th layers of the visual encoder for anomaly detection. We set the number of learnable tokens $M$ to 8 and used the linear layer as our adapter, setting the temperature parameter $\tau$ in Eq. (2) to 0.07. For the anomaly synthesis tasks, we uniformly sampled the intensity variation factor $\gamma$ in Eq. (4) within the range of $[-0.6, -0.4) \cup [0.4, 0.6)$, and $\alpha$ in Eq. (5) within $[\sqrt{2}, 4)$. For the CutPaste and GaussIntensityChange tasks, we utilized a Perlin noise generator~\cite{perlin1985image} to capture various anomalous shapes and binarize them into the anomaly mask $Y$. For the Source task, since it needs to compute the radius of the anomalous region in any direction, we used ellipses or rectangles with random sizes and rotation angles as the anomaly mask. We employed three types of anomaly synthesis tasks with equal probability for anomaly synthesis. The input image size was $224 \times 224$ pixels. For the three datasets, we used a consistent hyperparameter setting. Each experiment was independently conducted three times, sampling different support sets for MediCLIP training.

\noindent \textbf{Baselines and Metrics.} To comprehensively evaluate the few-shot anomaly detection performance of MediCLIP, we established the following baselines for comparison: 
deep feature embedding-based methods PatchCore~\cite{roth2022towards} and SimpleNet~\cite{liu2023simplenet}, reconstruction-based methods SQUID~\cite{xiang2023squid} and AnoDDPM~\cite{wyatt2022anoddpm}, self-supervised learning-based method CutPaste~\cite{li2021cutpaste}, one-class classification method PatchSVDD~\cite{yi2020patch}, and few-shot anomaly detection method RegAD~\cite{huang2022registration}. For the evaluation metrics, we used Image-AUROC to evaluate anomaly detection performance and Pixel-AUROC to evaluate anomaly localization performance.

\subsection{Experimental Results}

We compared the anomaly detection performance of MediCLIP with other methods on three datasets, as shown in Table \ref{tab:table1}. Under the few-shot setting, other methods struggle to achieve good generalization due to the scarcity of normal images. In contrast, through a series of carefully crafted anomaly synthesis tasks, MediCLIP effectively transfers the powerful generalization capability of the CLIP model to the task of medical anomaly detection. MediCLIP achieved a performance improvement of approximately 10\% over other methods. Table \ref{tab:table2} displays the anomaly localization performance of MediCLIP and other methods on the BUSI dataset~\cite{al2020dataset}, indicating that MediCLIP is more effective in locating lesion areas. Fig. \ref{fig:fig3} provides a visualization of the anomaly localization results by MediCLIP, demonstrating its accurate identification of anomaly regions, thereby offering valuable references for real medical decision-making. 

Table \ref{tab:table3} demonstrates MediCLIP's zero-shot anomaly detection performance on various datasets. The experimental results indicate that MediCLIP's anomaly detection capability can be generalized across different tasks, demonstrating that the multi-task anomaly synthesis strategy can effectively simulate common anomaly patterns contained in various medical images, further evidencing its potential as a unified solution in medical anomaly detection.

\begin{table}[t]
  \renewcommand\arraystretch{1.0}
  \centering
  \caption{Ablation study results for learnable token and adapter in MediCLIP with support set size $k = 16$, using Image-AUROC (\%) as the evaluation metric.}
    \resizebox{0.56\linewidth}{!}{
    \footnotesize
    \begin{tabular}{c|c|c|c|c|c}
    \bottomrule
          & \makecell{ {\;Learnable\;}\\ Token} & {\;Adapter\;} & {\;CheXpert\;} & {\;BrainMRI\;} & {\;\;\;\;BUSI\;\;\;\;} \\
    \hline
    {\quad 1 \quad}     & -     & -     & 42.5{\scriptsize$\pm$0.0} & 41.6{\scriptsize$\pm$0.0} & 52.6{\scriptsize$\pm$0.0} \\
    {\quad 2 \quad}     & $\checkmark$     & -     & 57.7{\scriptsize$\pm$1.3} & 71.8{\scriptsize$\pm$1.3} & 80.3{\scriptsize$\pm$1.4} \\
    {\quad 3 \quad}     & -     & $\checkmark$     & 71.4{\scriptsize$\pm$2.1} & 94.2{\scriptsize$\pm$0.9} & 90.3{\scriptsize$\pm$0.7} \\
    {\quad 4 \quad}     & $\checkmark$     & $\checkmark$     & \textbf{72.5{\scriptsize$\pm$1.7}} & \textbf{94.8{\scriptsize$\pm$0.5}} & \textbf{90.8{\scriptsize$\pm$0.4}} \\
    \toprule
    \end{tabular}%
    }
  \label{tab:table4}
\end{table}

\begin{table}[t]
  \renewcommand\arraystretch{1.0}
  \centering
  \caption{Anomaly detection performance of MediCLIP under different training settings, using Image-AUROC (\%) as the evaluation metric.}
    \resizebox{0.76\linewidth}{!}{
    \footnotesize
    \begin{tabular}{c|c|c|c|c}
    \bottomrule
    \multicolumn{2}{c|}{} & {\;\;CheXpert\;\;} & {\;\;BrainMRI\;\;} & {\;\;\;\;BUSI\;\;\;\;} \\
    \hline
    \multicolumn{1}{c|}{\multirow{2}{*}{\makecell{Auxiliary \\ Datasets}}} & MVTec-AD & 64.4{\scriptsize$\pm$0.7} & 88.2{\scriptsize$\pm$1.1} & 77.0{\scriptsize$\pm$0.9} \\
    \cline{2-5}          & VisA  & 55.1{\scriptsize$\pm$1.0} & 85.2{\scriptsize$\pm$1.1} & 74.7{\scriptsize$\pm$2.1} \\
    \hline
    \multicolumn{1}{c|}{\multirow{6}{*}{\makecell{Anomaly \\ \;Synthesis\; \\ ($k=16$)}}} & CutPaste & 67.6{\scriptsize$\pm$1.3} & 94.8{\scriptsize$\pm$1.2} & 89.3{\scriptsize$\pm$1.8} \\
    \cline{2-5}          & GaussIntensityChange & 70.0{\scriptsize$\pm$1.8} & 94.0{\scriptsize$\pm$0.2} & 90.0{\scriptsize$\pm$0.6} \\
    \cline{2-5}          & Source & 67.3{\scriptsize$\pm$1.1} & 92.2{\scriptsize$\pm$1.0} & 87.9{\scriptsize$\pm$1.0} \\
    \cline{2-5}          & \makecell{\;\;GaussIntensityChange \&  CutPaste\;\;} & 71.8{\scriptsize$\pm$0.4} & \textbf{95.2{\scriptsize$\pm$0.4}} & 90.1{\scriptsize$\pm$0.8} \\
    \cline{2-5}          & \makecell{GaussIntensityChange \& Source} & 71.1{\scriptsize$\pm$1.1} & 94.1{\scriptsize$\pm$0.5} & 90.4{\scriptsize$\pm$1.1} \\
    \cline{2-5}          & All Synthesis Tasks & \textbf{72.5{\scriptsize$\pm$1.7}} & 94.8{\scriptsize$\pm$0.5} & \textbf{90.8{\scriptsize$\pm$0.4}} \\
    \toprule
    \end{tabular}%
    }
  \label{tab:table5}
\end{table}

\subsection{Ablation Studies}

Table \ref{tab:table4} validates the effectiveness of the learnable token and adapter in MediCLIP. We replaced the learnable token with hard prompt templates provided in~\cite{jeong2023winclip}, such as \texttt{`A photo of the [CLS]'}. Additionally, we used average pooling along the channel instead of the adapter. The experimental results in Table \ref{tab:table4} demonstrate the irreplaceability of the learnable token and adapter. Table \ref{tab:table5} compares the anomaly detection performance of MediCLIP under different training settings. For the auxiliary datasets, we follow the settings in~\cite{zhou2023anomalyclip,chen2023clip,chen2023zero}, using the industrial anomaly detection datasets MVTec-AD~\cite{bergmann2019mvtec} and VisA~\cite{zou2022spot} for model training, and assess the anomaly detection performance in medical images. Compared to auxiliary datasets, the medical anomaly synthesis task achieved better generalization performance. Furthermore, the combination of multiple tasks can further enhance the anomaly detection performance of MediCLIP.

\section{Conclusion}

This paper first comprehensively explores the task of few-shot medical image anomaly detection, a field with broad applications yet insufficiently studied. We introduce MediCLIP, an adaptation of the CLIP model tailored for medical anomaly detection, integrating multi-task anomaly synthesis. Characterized by its efficiency and superior performance, MediCLIP demonstrates effective detection of various diseases across diverse medical imaging types. Our comprehensive experiments on three medical datasets underscore MediCLIP's capability in accurate anomaly detection and localization, establishing its promise as a unified and vital tool in medical diagnostics.

\vspace{0.5cm}
\noindent \textbf{Acknowledgements}. This work was supported in part by the National Natural Science Foundation of China under Grant 62177034 and Grant 61972046, the Beijing Natural Science Foundation under grant Z190020, and the Proof of Concept Program of Zhongguancun Science City and Peking University Third Hospital under Grant HDCXZHKC2022202.

\bibliographystyle{splncs04}
\bibliography{ref}

\clearpage
\renewcommand*{\thefigure}{S\arabic{figure}}
\renewcommand*{\thetable}{S\arabic{table}}
\renewcommand*{\theequation}{S\arabic{equation}}
\renewcommand\thesection{\Alph{section}}
\setcounter{table}{0}
\setcounter{figure}{0}
\setcounter{equation}{0}
\setcounter{section}{0}
\appendix
\sectionfont{\centering}
\thispagestyle{empty}
\section*{Appendix}
\begin{table}[h]
  \renewcommand\arraystretch{1.1}
  \centering
  \caption{List of class tokens.}
    \resizebox{0.8\linewidth}{!}{
    \begin{tabular}{c|c}
    \bottomrule
    \multicolumn{1}{c|}{Normal} & Anomaly \\
    \hline
    \hline
    \multicolumn{1}{c|}{\texttt{normal}} & \texttt{abnormal} \\
    \multicolumn{1}{c|}{\texttt{healthy}} & \texttt{disease} \\
    \multicolumn{1}{c|}{\texttt{negative}} & \texttt{lesion} \\
    \multicolumn{1}{c|}{\texttt{unremarkable}} & \texttt{positive} \\
    \multicolumn{1}{c|}{\texttt{clear}} & \texttt{symptomatic} \\
    \multicolumn{1}{c|}{\texttt{asymptomatic}} & \texttt{pathological} \\
    \multicolumn{1}{c|}{\texttt{normal findings}} & \texttt{impaired} \\
    \multicolumn{1}{c|}{\texttt{no findings}} & \texttt{evidence of disease} \\
    \multicolumn{1}{c|}{\texttt{in good health}} & \texttt{abnormal finding} \\
    \multicolumn{1}{c|}{\;\;\texttt{no evidence of disease}\;\;} & \texttt{pathological condition} \\
          & {\;\;\texttt{pathological abnormality}\;\;} \\
    \toprule
    \end{tabular}%
    }
  \label{tab:tables1}
\end{table}

\end{document}